\documentclass[10pt,twocolumn,letterpaper]{article}

\usepackage{iccv}
\usepackage{times}
\usepackage{epsfig}
\usepackage{graphicx}
\usepackage{amsmath}
\usepackage{bm}
\usepackage{amssymb}
\usepackage{booktabs}
\usepackage{multirow}
\usepackage{adjustbox}
\usepackage[utf8x]{inputenc} 
\usepackage{csquotes}
\usepackage[table,xcdraw]{xcolor}
\usepackage{nicefrac}       

\usepackage[pagebackref=true,breaklinks=true,colorlinks,bookmarks=false,citecolor=blue,linkcolor=blue,urlcolor=blue]{hyperref}

\iccvfinalcopy 

\def\iccvPaperID{2617} 

\ificcvfinal\fi

\definecolor{mygreen}{rgb}{0.4, 0.69, 0.2}

\begin{document}

\title{Discriminative Region-based Multi-Label Zero-Shot Learning}

\author{Sanath Narayan\thanks{Equal contribution}~~$^{1}$ \quad Akshita Gupta\footnotemark[1]~~$^{1}$ \quad Salman Khan$^{2}$ \quad 
Fahad Shahbaz Khan$^{2,3}$ \\
\quad Ling Shao$^{1}$  \quad Mubarak Shah$^{4}$ \\
$^1$Inception Institute of Artificial Intelligence, UAE \quad $^2$Mohamed Bin Zayed University of AI, UAE \\
$^3$Linköping University, Sweden \quad
$^4$University of Central Florida, USA
}

\maketitle
\ificcvfinal\thispagestyle{empty}\fi

\begin{abstract}
Multi-label zero-shot learning (ZSL) is a more realistic counter-part of standard single-label ZSL since several objects can co-exist in a natural image. 
However, the occurrence of multiple objects complicates the reasoning and requires region-specific processing of visual features to preserve their contextual cues.
We note that the best existing multi-label ZSL method takes a shared approach towards attending to region features with a common set of attention maps for all the classes.
Such shared maps lead to diffused attention, which does not discriminatively focus on relevant locations when the number of classes are large. Moreover, mapping spatially-pooled visual features to the class semantics leads to inter-class feature entanglement, thus hampering the classification. Here, we propose an alternate approach towards region-based discriminability-preserving multi-label zero-shot classification. Our approach maintains the spatial resolution to preserve region-level characteristics and utilizes a bi-level attention module (BiAM) to enrich the features by incorporating both region and scene context information. The enriched region-level features are then mapped to the class semantics and only their class predictions are spatially pooled to obtain image-level predictions, thereby keeping the multi-class features disentangled. Our approach sets a new state of the art on two large-scale multi-label zero-shot benchmarks: NUS-WIDE and Open Images. On NUS-WIDE, our approach achieves an absolute gain of 6.9\% mAP for ZSL, compared to the best published results. Source code is available at \href{https://github.com/akshitac8/BiAM}{https://github.com/akshitac8/BiAM}.
\end{abstract}

\section{Introduction}
\begin{figure*}
    \centering
    \includegraphics[width=1\textwidth]{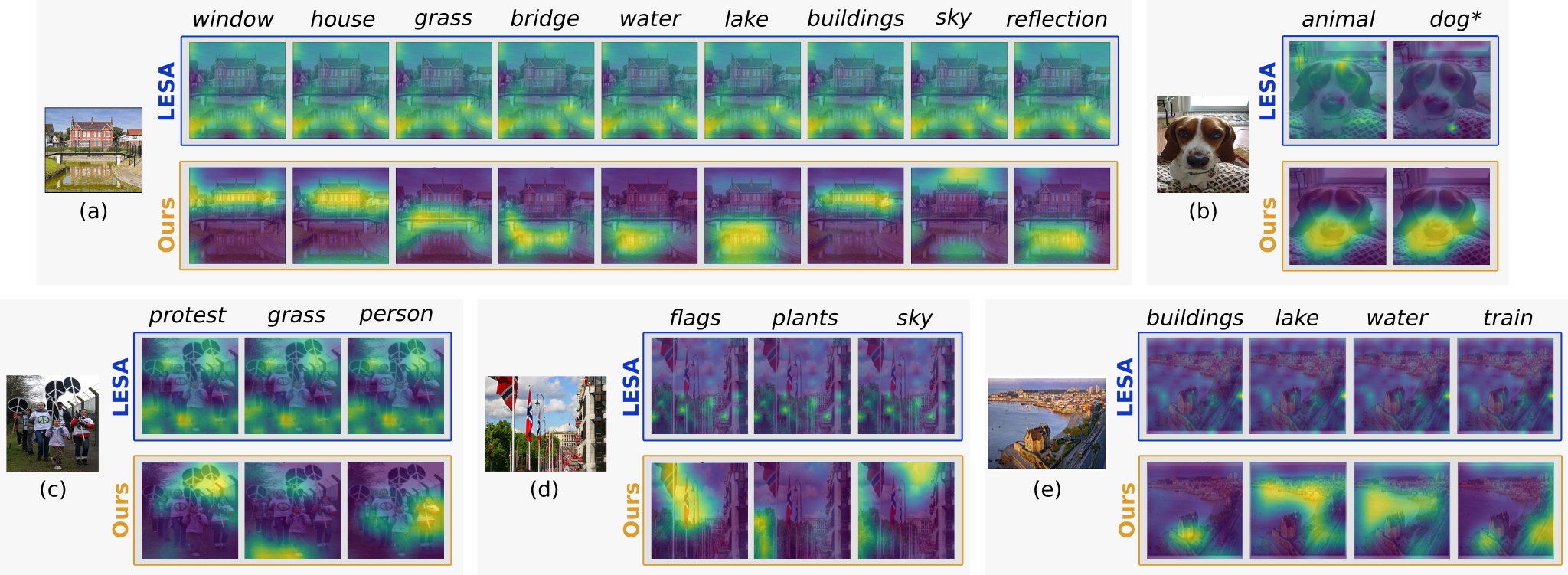}
    \caption{\textbf{Comparison, in terms of attention visualization, between shared attention-based LESA~\cite{huynh2020shared} and our approach on example NUS-WIDE test images}. For each image, the visualization of attentions of positive labels within that image are shown for LESA (top row) and our approach (bottom row). In the case of LESA, all classes in these examples are inferred from the eighth shared attention module except for \textit{dog} class in (b), which is inferred from the ninth module. As seen in these examples, these dominant attention maps struggle to discriminatively focus on relevant (class-specific) regions. In contrast, our proposed approach based on a bi-level attention module (BiAM) produces attention maps by preserving class-specific discriminability, leading to an enriched feature representation. Our BiAM effectively captures region-level semantics as well as global scene-level context, thereby enabling it to accurately attend to object class (\eg, \textit{window} class in (a)) \textit{and} abstract concepts (\eg, \textit{reflection} class in (a)). Best viewed zoomed in. \vspace{-0.15cm} }
    \label{fig:intro}
\end{figure*}

Multi-label classification strives to recognize all the categories (labels) present in an image. In the standard multi-label classification~\cite{wang2016cnn,yazici2020orderless,kipf2016semi,chen2019multi,nam2017maximizing,yeattention,you2020cross} setting, the category labels in both the train and test sets are identical. In contrast, the task of multi-label zero-shot learning (ZSL) is to recognize multiple new unseen categories in images at test time, without having seen the corresponding visual examples during training. In the generalized ZSL (GZSL) setting, test images can simultaneously contain multiple seen and unseen classes. GZSL is particularly challenging in the large-scale multi-label setting, where several diverse categories occur in an image (\eg, maximum of $117$ labels per image in NUS-WIDE~\cite{nuswide}) along with a large number of unseen categories at test time (\eg, $400$ unseen classes in Open Images~\cite{openimages}). Here, we investigate this challenging problem of multi-label (generalized) zero-shot classification.\\
\indent Existing multi-label (G)ZSL methods tackle the problem by using global image features~\cite{mensink2014costa,zhang2016fast}, structured knowledge graph~\cite{lee2018multi} and attention schemes~\cite{huynh2020shared}. Among these, the recently introduced LESA~\cite{huynh2020shared} proposes a shared attention scheme based on region-based feature representations and achieves state-of-the-art results. LESA learns multiple attention maps that are shared across all categories. The region-based image features are weighted by these shared attentions and then spatially aggregated. Subsequently, the aggregated features are projected to the label space via a joint visual-semantic embedding space.\\
\indent While achieving promising results, LESA suffers from two key limitations. Firstly, classification is performed on features obtained using a set of attention maps that are shared across all the classes.
In such a shared attention framework, many categories are observed to be inferred from only a few dominant attention maps, which tend to be diffused across an image rather than
discriminatively focusing on regions likely belonging to a specific class (see Fig.~\ref{fig:intro}).
This is problematic for large-scale benchmarks comprising several hundred categories, \eg, more than 7$k$ seen classes in Open Images~\cite{openimages} with significant inter and intra-class variations. Secondly, attended features are spatially pooled before projection to the label space, thus entangling the multi-label information in the collapsed image-level feature vectors. Since multiple diverse labels can appear in an image, the class-specific discriminability within such a collapsed representation is severely hampered. 

\subsection{Contributions}
To address the aforementioned problems, we pose large-scale multi-label ZSL as a region-level classification problem.
We introduce a simple yet effective region-level classification framework that maintains the spatial resolution of features to keep the multi-class information disentangled for dealing with large number of co-existing classes in an image.
Our framework comprises a bi-level attention module (BiAM) to contextualize and obtain highly discriminative region-level feature representations.
Our BiAM contains region and global (scene) contextualized blocks and enables reasoning about all the regions together using pair-wise relations between them, in addition to utilizing the holistic scene context. The region contextualized block enriches each region feature by attending to all regions within the image whereas the scene contextualized block enhances the region features based on their congruence to the scene feature representation. The resulting discriminative features, obtained through our BiAM, are then utilized to perform region-based classification through a compatibility function. Afterwards, a spatial \textit{top}-$k$ pooling is performed over each class to obtain the final predictions.
Experiments are performed on two challenging large-scale multi-label zero-shot benchmarks:  NUS-WIDE~\cite{nuswide} and Open Images~\cite{openimages}. Our approach performs favorably against existing methods, setting a new state of the art on both benchmarks. Particularly, on NUS-WIDE, our approach achieves an absolute gain of $6.9\%$ in terms of mAP for the ZSL task, over the best published results~\cite{huynh2020shared}.

\section{Proposed Method}

Here, we introduce a region-based discriminability-preserving multi-label zero-shot classification framework aided by learning rich features that explicitly encodes both region as well as global scene contexts in an image.

\noindent\textbf{Problem Formulation:}
Let $\mathbf{x}\!\in\!\mathcal{X}$ denote the feature instances of a multi-label image $i \!\in\! \mathcal{I}$ and $\mathbf{y} \!\in\! \{0,1\}^S$ the corresponding multi-hot labels from the set of $S$ seen class labels $\mathcal{C}^s$. Further, let $\mathbf{A}_S {\in} \mathbb{R}^{S\times {d_a}}$ denote the $d_a$-dimensional attribute embeddings, which encode the semantic relationships between $S$ seen classes. With $n_p$ as the number of  positive labels in an image, we denote the set of attribute embeddings for the image as $a_\mathbf{y} {=} \{\mathbf{A}_j, \forall j\!:\!\mathbf{y}[j]{=}1 \}$, where $|a_\mathbf{y}|{=}n_p$.
The goal in (generalized) zero-shot learning is to learn a mapping $f(\mathbf{x})\!:\!\mathcal{X} {\rightarrow} \{0,1\}^S$ aided by the attribute embeddings $a_\mathbf{y}$, such that the mapping can be adapted to include the $U$ unseen classes (with embeddings $\mathbf{A}_U {\in} \mathbb{R}^{U\times {d_a}}$) at test time, \ie, $f(\mathbf{x})\!:\!\mathcal{X} {\rightarrow} \{0,1\}^U$ for ZSL and $f(\mathbf{x})\!:\!\mathcal{X} {\rightarrow} \{0,1\}^C$ for the GZSL setting. Here, $C{=}S+U$ represents the total number of seen and unseen classes.

\begin{figure}[t]
    \centering
    \includegraphics[width=0.95\columnwidth]{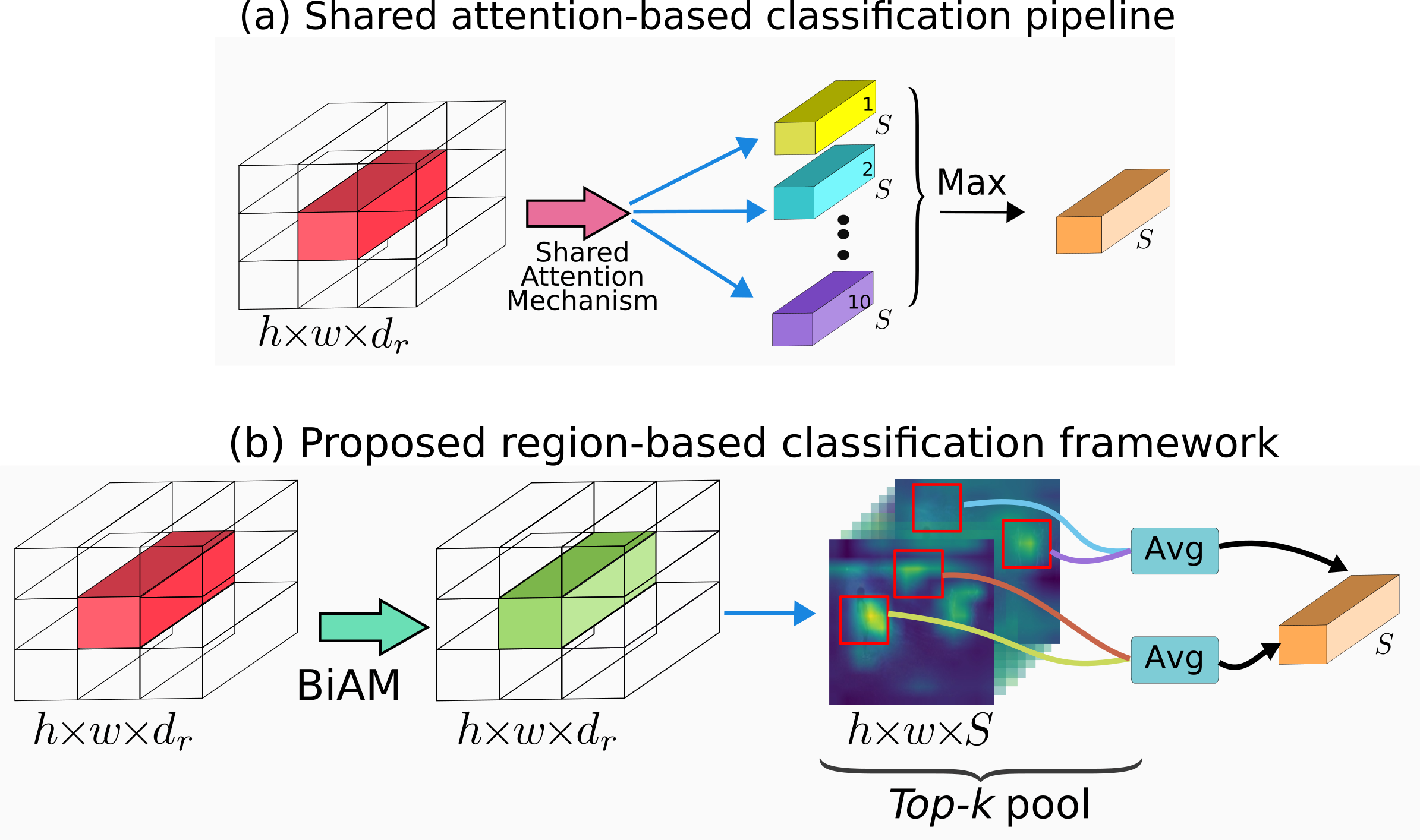}
    \caption{\textbf{Comparison of our region-level classification framework (b) with the shared attention-based classification pipeline (a) in~\cite{huynh2020shared}}. The shared attention-based pipeline performs an attention-weighted spatial averaging of the region-based features to generate a feature vector per shared attention. These (spatially pooled) features are then classified to obtain $S$ class scores per shared attention, which are max-pooled to obtain image-level class predictions. In contrast, our framework minimizes inter-class feature entanglement by enhancing the region-based features through a feature enrichment mechanism, which preserves the spatial resolution of the features. Each region-based enriched feature representation is then classified to $S$ seen classes. Afterwards, \textit{per class} \textit{top}-$k$ activations are aggregated to obtain image-level predictions. \vspace{-0.5cm}
    }
    \label{fig:classify_compare}
\end{figure}

\subsection{Region-level Multi-label ZSL\label{sec:regionwise}}

\begin{figure*}
    \centering
    \includegraphics[width=0.96\textwidth]{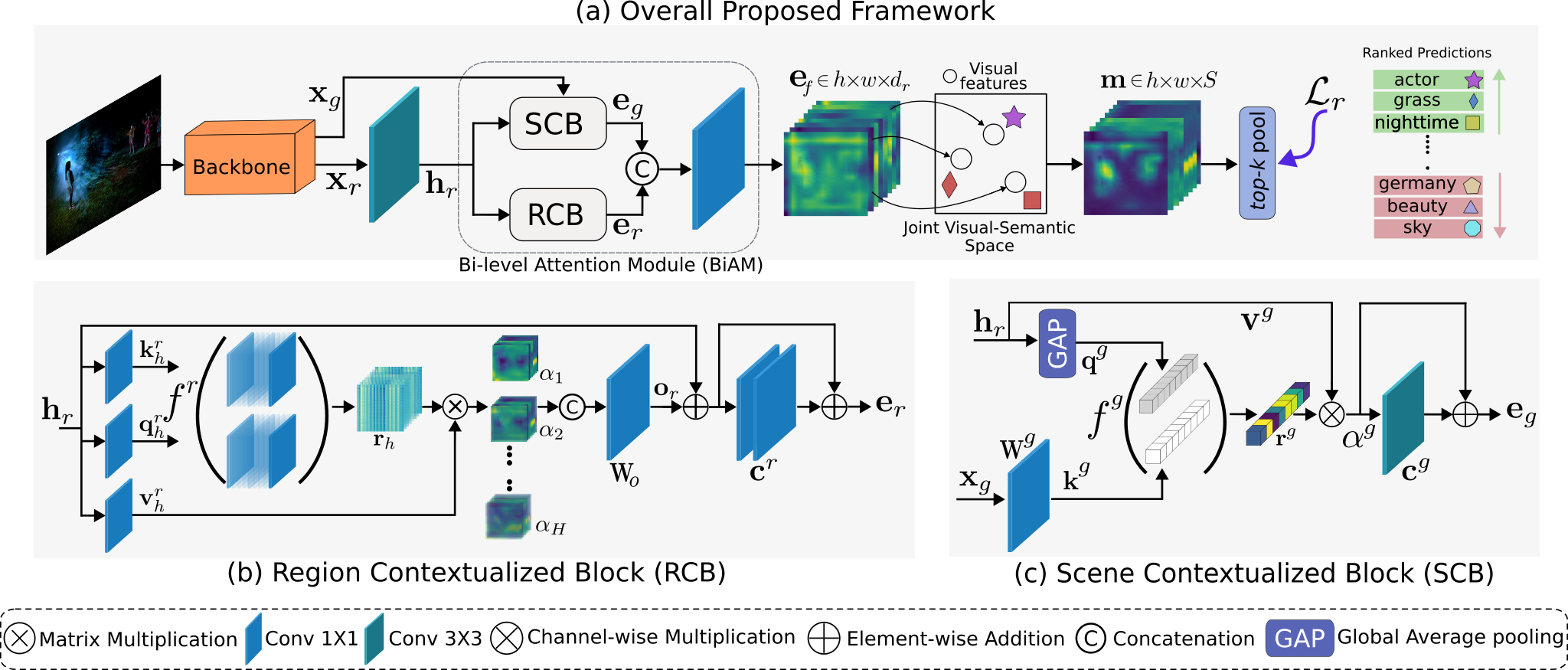}
    \caption{\textbf{Our region-level multi-label (G)ZSL framework:} The top row shows an overview of our network architecture. Given an image, the region-level features $\mathbf{x}_r$ are first obtained using a backbone. The region features are enriched using a Bi-level Attention Module (BiAM). This module incorporates region (b) and scene (c) contextualized blocks which learn to aggregate region-level and scene-specific context, respectively, which is in turn used to enhance the region features. The enriched features $\mathbf{e}_f$ are mapped to the joint visual-semantic space to relate them with class semantics, obtaining $\mathbf{m}$. Per-class region-based prediction scores are then spatially pooled to generate final image-level predictions. Notably, our design ensures \emph{region-level feature enrichment} while preserving the spatial resolution uptil class predictions are made, which \emph{minimizes inter-class feature entanglement}, a key requisite for large-scale multi-label (G)ZSL. \vspace{-0.2cm}
    }
    \label{fig:overall_arch}
\end{figure*}

As discussed earlier, recognizing diverse and wide range of category labels in images under the (generalized) zero-shot setting is challenging.
The problem arises, primarily, due to the entanglement of features of the various different classes present in an image. 
Fig.~\ref{fig:classify_compare}(a) illustrates this feature entanglement in the shared attention-based classification pipeline~\cite{huynh2020shared} that integrates multi-label features by performing a weighted spatial averaging of the region-based features based on the shared-attention maps. In this work, we argue that entangled feature representations are sub-optimal for multi-label classification and instead propose to alleviate this issue by posing large-scale multi-label ZSL as a region-level classification problem. To this end, we introduce a simple but effective region-level classification framework that first enriches the region-based features by the proposed feature enrichment mechanism. It then classifies the enriched region-based features followed by spatially pooling the \textit{per-class} region-based scores to obtain the final image-level class predictions (see Fig.~\ref{fig:classify_compare}(b)). Consequently, our framework minimizes inter-class feature entanglement and enhances the classification performance.
Fig.~\ref{fig:overall_arch} shows our overall proposed framework. Let $\mathbf{e}_f \in \mathbb{R}^{h\times w\times d_r}$ be the output region-based features, which are to be classified, from our proposed enrichment mechanism (\ie, BiAM). Here, $h,w$ denote the spatial extent of the region-based features with $h\cdot w$ regions. These features $\mathbf{e}_f$ are first aligned with the class-specific attribute embeddings of the seen classes. This alignment is performed, \ie, a joint visual-semantic space is learned, so that the classifier can be adapted to the unseen classes at test time. The aligned region-based features are classified to obtain class-specific response maps $\mathbf{m}\in\mathbb{R}^{h\times w\times S}$ given by,
\begin{equation}
    \label{eq:response_maps}
    \mathbf{m} = \mathbf{e}_f\mathbf{W}_a\mathbf{A}_S^\top, \qquad \text{s.t., } \mathbf{A}_S \in \mathbb{R}^{S\times d_a},
\end{equation}
where $\mathbf{W}_a \in \mathbb{R}^{d_r\times d_a}$ is a learnable weight matrix that is used to reshape the visual features to attribute embeddings of seen classes ($\mathbf{A}_S$). The response maps are then \textit{top}-$k$ pooled along the spatial dimensions to obtain image-level  \textit{per-class} scores $\mathbf{s} \in \mathbb{R}^S$,
which are then utilized for training the network (in Sec.~\ref{sec:training_loss}). Such a region-level classification, followed by a score-level pooling, helps to preserve the discriminability of the features in each of the $h\cdot w$ regions by minimizing the feature entanglement of different positive classes occurring in the image. 

The aforementioned region-level multi-label ZSL framework relies on discriminative region-based features. Standard region-based features $\mathbf{x}$ only encode local region-specific information and do not explicitly reason about all the regions together.
Moreover, region-based features do not possess image-level holistic scene information. 
Next, we introduce a bi-level attention module (BiAM) to enhance feature discriminability and generate enriched features $\mathbf{e}_f$.
\subsection{Bi-level Attention Module}
Here, we present a bi-level attention module (BiAM) that enhances region-based features by incorporating both region and scene context information, without sacrificing the spatial resolution. Our BiAM comprises region and scene contextualized blocks, which are described next.
\subsubsection{Region Contextualized Block}
The region-contextualized block (RCB) enriches the region-based latent features $\mathbf{h}_r$ by capturing the contexts from different regions in the image. We observe encoding the individual contexts of different regions in an image to improve the discriminability of standard region-based features, \eg, the context of a region with \textit{window} can aid in identifying other possibly texture-less regions in the image as \textit{house} or \textit{building}. Thus, inspired by the multi-headed self-attention \cite{vaswani2017attention}, our RCB allows the features in different regions to interact with each other and identify the regions to be paid more attention to for enriching themselves (see Fig.~\ref{fig:overall_arch}(b)). To this end, the input features $\mathbf{x}_r\in\mathbb{R}^{h\times w\times d_r}$ are first processed by a $3{\times} 3$ convolution layer to obtain latent features $\mathbf{h}_r \in \mathbb{R}^{h\times w \times d_r}$.
These latent features are then projected to a low-dimensional space ($d_r' = \nicefrac{d_r}{H}$) to create query-key-value triplets using a total of $H$ projection heads,
\begin{equation}
 \mathbf{q}_h^r = \mathbf{h}_r\mathbf{W}_h^Q, \quad \mathbf{k}_h^r =  \mathbf{h}_r\mathbf{W}_h^K, \quad \mathbf{v}_h^r = \mathbf{h}_r\mathbf{W}_h^V,    
\end{equation}
where $h {\in} \{1,2, .. , H\}$ and $\mathbf{W}_h^Q , \mathbf{W}_h^K , \mathbf{W}_h^V$ are learnable weights of $1{\times}1$ convolution layers with input and output channels as $d_r$ and $d_r'$, respectively. The \emph{query} vector (of length $d_r'$) derived from each region feature\footnote{\label{footnote:hw_regions}Query $\mathbf{q}_h^r \in \mathbb{R}^{h{\times} w{\times} d_r'}$ can be considered as $h\cdot w$ queries represented by $d_r'$ features each. Similar observation holds for keys, values, \etc.} is used to find its correlation with the \emph{keys} obtained from all the region features, while the \textit{value} embedding holds the status of the current form of each region feature. 

Given these triplets for each head, first, an intra-head processing is performed by relating each query vector with `\emph{keys}' derived from the $h\cdot w$ region features. The resulting normalized relation scores ($\mathbf{r}_h \in \mathbb{R}^{hw \times hw}$) from the softmax function ($\sigma$) are used to reweight the corresponding `\emph{value}' vectors. {Without loss of generality}\textsuperscript{\ref{footnote:hw_regions}}, the attended features ${\bm\alpha}_h  \in \mathbb{R}^{h\times w \times d_r'}$ are given by, 
\begin{equation}
     {\bm\alpha}_h = \mathbf{r}_h \mathbf{v}_h^r , \quad \text{ where }\; \mathbf{r}_h = \sigma ( \frac{\mathbf{q}_h^r \mathbf{k}^{r\top}_h }{\sqrt{d_r'}} ).
\end{equation}
Next, these low-dimensional {self-attended} features from each head are channel-wise concatenated and processed by a convolution layer $\mathbf{W}_o$ to generate output $\mathbf{o}_r \in \mathbb{R}^{h {\times} w {\times} d_r}$, 
\begin{equation}
    \mathbf{o}_r = [\bm\alpha_1; \bm\alpha_2; \ldots \bm\alpha_H]\mathbf{W}_o.
\end{equation}
To encourage the network to selectively focus on adding complimentary information to the `source' latent feature $\mathbf{h}_r$, a residual branch is added to the attended features $\mathbf{o}_r$ and further processed with a small residual sub-network $c^r(\cdot)$, comprising two $1{\times}1$ convolution layers, to help the network first focus on the local neighbourhood and then progressively pay attention to the other-level features. The enriched region-based features $\mathbf{e}_r \in \mathbb{R}^{h {\times} w {\times} d_r}$ from the RCB are given by,
\begin{equation}
\mathbf{e}_r = c^r(\mathbf{h}_r + \mathbf{o}_r) + (\mathbf{h}_r + \mathbf{o}_r).
\end{equation}
Consequently, the discriminability of the latent features $\mathbf{h}_r$ is enhanced by self-attending to the context of different regions in the image, resulting in enriched features $\mathbf{e}_r$.

\subsubsection{Scene Contextualized Block}
As discussed earlier, the RCB captures the regional context in the image, enabling reasoning about all regions together using pair-wise relations between them. In this way, RCB enriches the latent feature inputs $\mathbf{h}_r$. However, such a region-based contextual attention does not effectively encode the global scene-level context of the image, which is necessary for understanding abstract scene concepts like \textit{night-time}, \textit{protest}, \textit{clouds}, \etc. Understanding such labels from local regional contexts is challenging due to their abstract nature.
Thus, in order to better capture the holistic scene-level context, we introduce a scene contextualized block (SCB) within our BiAM. 
Our SCB attends to the region-based latent features $\mathbf{h}_r$, based on their congruence with the global image feature $\mathbf{x}_g$ (see Fig.~\ref{fig:overall_arch}(c)). To this end, the learnable weights $\mathbf{W}^g$ project the features $\mathbf{x}_g$ to a $d_r$-dimensional space to obtain the global `\textit{key}' vectors $\mathbf{k}^g \in \mathbb{R}^{d_r}$, while the latent features $\mathbf{h}_r$ are spatially average pooled to create the `\textit{query}' vectors $\mathbf{q}^g \in \mathbb{R}^{d_r}$, 
\begin{equation}
    \mathbf{q}^g = \text{GAP}(\mathbf{h}_r), \quad \mathbf{k}^g =  \mathbf{x}_g \mathbf{W}^g, \quad \mathbf{v}^g = \mathbf{h}_r.
\end{equation}
The region-based latent features $\mathbf{h}_r$ are retained as `\textit{value}' features $\mathbf{v}^g$. Given these query-key-value triplets, first, the \textit{query} $\mathbf{q}^g$ is used to find its correlation with the \textit{key} $\mathbf{k}^g$. The resulting relation score vectors $\mathbf{r}^g\in\mathbb{R}^{d_r}$ are then used to reweight the corresponding channels in \textit{value} features to obtain the attended features ${\bm\alpha}^g  \in \mathbb{R}^{h\times w \times d_r}$, given by,
\begin{equation}
     {\bm\alpha}^g = \mathbf{v}^g \otimes \mathbf{r}^g, \quad \text{ where }\; \mathbf{r}^g = \text{sigmoid} ( {\mathbf{q}^g * \mathbf{k}^g } ),
\end{equation}
where $\otimes$ and $*$ denote channel-wise and element-wise multiplications. The channel-wise operation is chosen here since we want to use the global contextualized features to dictate kernel-wise importance of the feature channels for aggregating relevant contextual cues without disrupting the local filter signature.
Similar to RCB, to encourage the network to selectively focus on adding complimentary information to the `source' $\mathbf{h}_r$, a residual branch is added after processing the attended features through a $3{\times}3$ convolution layer $c^g(\cdot)$. The scene-context enriched features $\mathbf{e}_g \in \mathbb{R}^{h {\times} w {\times} d_r}$ from the SCB are given by,
\begin{equation}
\mathbf{e}_g = c^g({\bm\alpha}^g) + \mathbf{h}_r.
\end{equation}
\begin{figure}[t]
    \centering
    \includegraphics[width=\columnwidth]{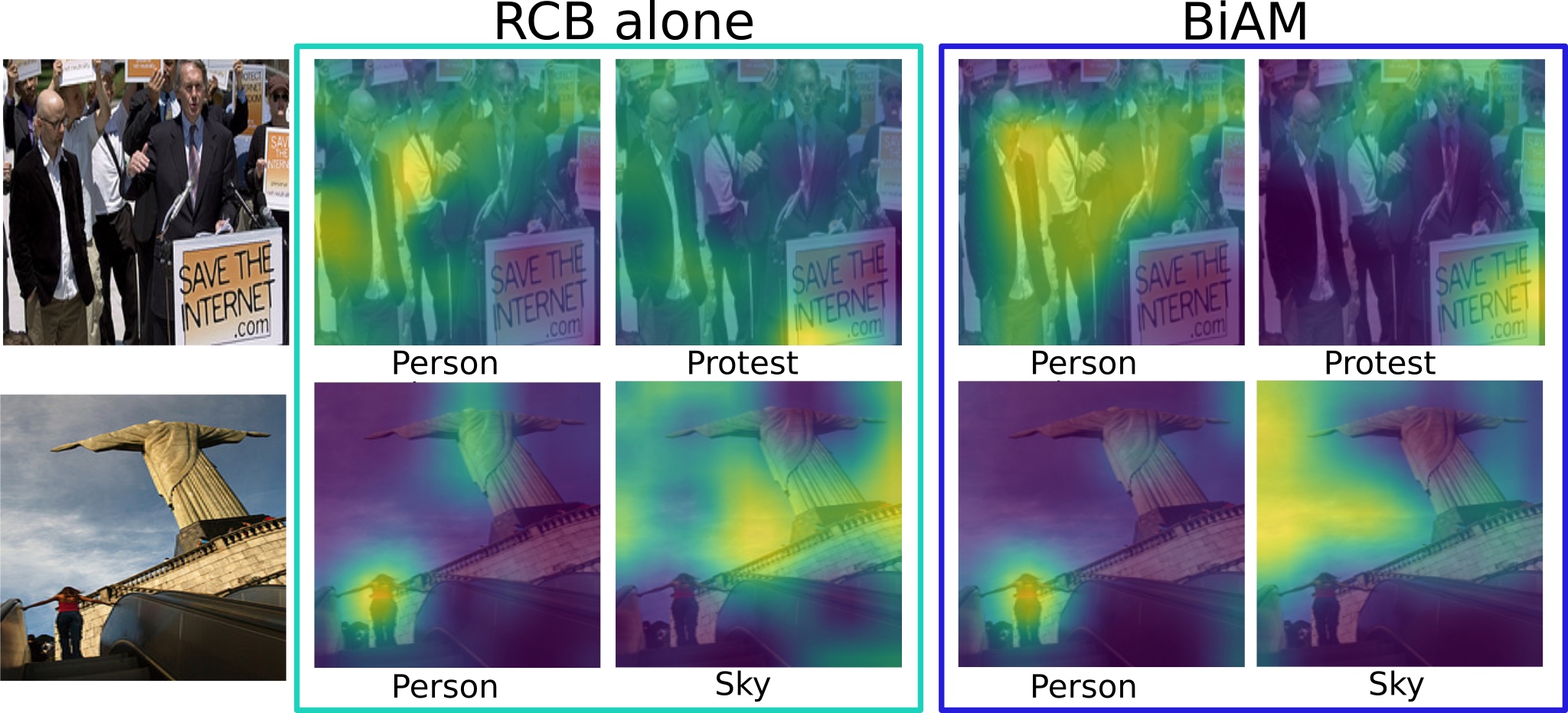}
    \caption{\textbf{Effect of enhancing the region-based features through our feature enrichment mechanism: BiAM}. The two complementary RCB and SCB blocks in BiAM integrate region-level semantics and global scene-level context, leading to a more discriminative feature representation. While RCB alone (on the left) is able to capture the region-level semantics of \textit{person} class, it confuses those related to \textit{protest} label. However, encoding the global scene-level context from the SCB in BiAM (on the right) improves the semantic recognition of scene-level concepts like \textit{protest}.\vspace{-0.2cm}}
    \label{fig:qual_rcb_scb}
\end{figure}
In order to ensure the enrichment due to both region and global contexts are well captured, the enriched features ($\mathbf{e}_r$ and  $\mathbf{e}_g$) from both region and scene contextualized blocks are {channel-wise} concatenated and processed through a $1\times1$ {channel-reducing} convolution layer $c^f(\cdot)$ to obtain the final enriched features $\mathbf{e}_f \in \mathbb{R}^{h {\times} w {\times} d_r}$, given by,
\begin{equation}
  \mathbf{e}_f = c^f([\mathbf{e}_r ; \mathbf{e}_g]).
\end{equation}
Fig.~\ref{fig:qual_rcb_scb} shows that encoding scene context into the region-based features improves the attention maps of scene level labels (\eg, \textit{protest}), which were hard to attend to using only the region context. Consequently, our bi-level attention module effectively reasons about all the image regions together using pair-wise relations between them, while being able to utilize the whole image (holistic) scene as context.

\subsection{Training and Inference\label{sec:training_loss}}
As discussed earlier, discriminative region-based features $\mathbf{e}_f$ are learned and region-wise classified to obtain class-specific response maps $\mathbf{m} \in \mathbb{R}^{h {\times} w {\times} S}$ (using Eq.~\ref{eq:response_maps}). The response maps $\mathbf{m}$ are further \textit{top}-$k$ pooled spatially to compute the image-level \textit{per-class} scores $\mathbf{s}\in\mathbb{R}^S$. The network is trained using a simple, yet effective ranking loss $\mathcal{L}_{rank}$ on the predicted scores $\mathbf{s}$, given by,
\begin{equation}
\label{eq:ranking_loss}
    \mathcal{L}_{rank} = \sum_i \sum_{p\in \mathbf{y}_p, n\notin \mathbf{y}_p} \max(\mathbf{s}_i[n] - \mathbf{s}_i[p] + 1, 0).
\end{equation}
Here, $\mathbf{y}_p\!=\!\{j\!:\!\mathbf{y}[j]{=}1\}$ denotes the positive labels in image $i$. The ranking loss ensures that the predicted scores of the positive labels present in the image rank ahead, by a margin of at least $1$, of the negative label scores.

At test time, for the multi-label ZSL task, the unseen class attribute embeddings $\mathbf{A}_U \!\in\! \mathbb{R}^{U\times d_a}$ of the respective unseen classes are used (in place of $\mathbf{A}_S$) for computing the class-specific response maps $\mathbf{m} \in \mathbb{R}^{h {\times} w {\times} U}$ in Eq.~\ref{eq:response_maps}. As in training, these response maps are then \textit{top}-$k$ pooled spatially to compute the image-level \textit{per-class} scores $\mathbf{s}\in\mathbb{R}^U$. Similarly, for the multi-label GZSL task, the concatenated embeddings ($\mathbf{A}_C {\in} \mathbb{R}^{C\times {d_a}}$) of all the classes $C=S+U$ are used to classify the multi-label images.

\section{Experiments\label{sec:experiments}}
\noindent\textbf{Datasets:} We evaluate our approach on two benchmarks: NUS-WIDE~\cite{nuswide} and Open Images~\cite{openimages}.
The \textbf{NUS-WIDE} dataset comprises nearly $270$K images with $81$ human-annotated categories, in addition to the $925$ labels obtained from Flickr user tags. As in~\cite{huynh2020shared,zhang2016fast}, the $925$ and $81$ labels are used as seen and unseen classes, respectively. 
The \textbf{Open Images} (v4) is a large-scale dataset comprising nearly $9$ million training images along with $41{,}620$ and $125{,}456$ images in validation and test sets. 
It has annotations with human and machine-generated labels. Here, $7{,}186$ labels, with at least $100$ training images, are selected as seen classes. The most frequent $400$ test labels that are absent in the training data are selected as unseen classes, as in \cite{huynh2020shared}. \\
\noindent\textbf{Evaluation Metrics:}
We use F1 score at \textit{top}-$K$ predictions and mean Average Precision (mAP) as evaluation metrics, as in~\cite{veit2017learning,huynh2020shared}. The model's ability to correctly rank labels in each image is measured by the F1, while the its image ranking accuracy for each label is captured by the mAP.\\
\noindent \textbf{Implementation Details:} 
Pretrained VGG-19~\cite{simonyan2014very} is used to extract features from multi-label images, as in~\cite{zhang2016fast,huynh2020shared}.
The region-based features (of size $h,w{=}14$ and $d_r{=}512$) from \textit{Conv}${}_5$ are extracted along with the global features of size $d_g{=}4{,}096$ from FC$7$. As in~\cite{huynh2020shared}, $\ell_2$-normalized $300$-dimensional GloVe~\cite{pennington2014glove} vectors of the class names are used as the attribute embeddings $\mathbf{A}_C$. The two $3{\times}3$ convolutions (input and output channels are set to $512$) are followed by ReLU and batch normalization layers. The $k$ for \textit{top}-$k$ pooling is set to $10$, while the heads $H{=}8$. For training, we use the ADAM optimizer with ($\beta_1,\beta_2$) as ($0.5$, $0.999$) and a gradual warm-up learning rate scheduler with an initial $lr$ of $1e^{-3}$. Our model is trained with a mini-batch size of $32$ for $40$ epochs on NUS-WIDE and $2$ epochs on Open Images.

\subsection{State-of-the-art Comparison}
\noindent\textbf{NUS-WIDE:} 
The state-of-the-art comparison for zero-shot (ZSL) and generalized zero-shot (GZSL) classification is presented in Tab.~\ref{tab:sota_nuswide}. The results are reported in terms of mAP and F1 score at \textit{top}-$K$ predictions ($K {\in} \{3,5\}$). The approach of Fast0Tag~\cite{zhang2016fast}, which finds principal directions in the attribute embedding space for ranking the positive tags ahead of negative tags, achieves $15.1$ mAP on the ZSL task. The recently introduced LESA~\cite{huynh2020shared}, which employs a shared multi-attention mechanism to recognize labels in an image, improves the performance over Fast0Tag, achieving $19.4$ mAP. Our approach outperforms LESA with an absolute gain of $6.9\%$ mAP. Furthermore, our approach achieves consistent improvement over the state-of-the-art in terms of F1 ($K{\in}\{3,5\}$), achieving gains as high as $2.0\%$ at $K{=}5$.

Similarly, on the GZSL task, our approach achieves an mAP score of $9.3$, outperforming LESA with an absolute gain of $3.7\%$. Moreover, consistent performance improvement in terms of F1 is achieved over LESA by our approach, with absolute gains of $1.5\%$ and $2.2\%$ at $K{=}3$ and $K{=}5$.

\begin{table}[t]
\centering
\caption{\textbf{State-of-the-art comparison for multi-label  ZSL and GZSL tasks on NUS-WIDE}. We report the results in terms of mAP and F1 score at $K{\in}\{3,5\}$. Our approach outperforms the state-of-the-art for both ZSL and GZSL tasks, in terms of mAP and F1 score. Best results are in bold.}
\adjustbox{width=\linewidth}{
\begin{tabular}{ccccc} 
\toprule[0.15em]
\multirow{1}{*}{\textbf{ Method}} & \multirow{1}{*}{\textbf{Task }} & \multirow{1}{*}{\textbf{mAP }} & \multicolumn{1}{c}{\cellcolor[HTML]{EEEEEE}\textbf{F1 (K = 3) }} & \multicolumn{1}{c}{\cellcolor[HTML]{DAE8FC}\textbf{F1 (K = 5) }}   \\
\toprule[0.15em]
\multirow{2}{*}{CONSE~\cite{norouzi2013zero}} & ZSL & 9.4 & 21.6 & 20.2   \\
 & GZSL & 2.1 &  7.0 & 8.1  \\ 
\cmidrule(lr){2-5}
\multirow{2}{*}{LabelEM~\cite{akata2015label}} & ZSL & 7.1 & 19.2  & 19.5   \\
 & GZSL & 2.2 &  9.5  & 11.3   \\ 
\cmidrule(lr){2-5}
\multirow{2}{*}{Fast0Tag~\cite{zhang2016fast}} & ZSL & 15.1 & 27.8 & 26.4  \\
 & GZSL & 3.7 & 11.5 & 13.5  \\ 
\cmidrule(lr){2-5}
\multirow{2}{*}{Attention per Label~\cite{kim2018bilinear}} & ZSL  & 10.4 & 25.8 & 23.6 \\
 & GZSL & 3.7 &  10.9 & 13.2   \\ 
\cmidrule(lr){2-5}
\multirow{2}{*}{Attention per Cluster~\cite{huynh2020shared}} & ZSL & 12.9 & 24.6 &  22.9  \\
 & GZSL  & 2.6 &  6.4 & 7.7  \\ 
\cmidrule(lr){2-5}
\multirow{2}{*}{LESA~\cite{huynh2020shared}} & ZSL & 19.4 & 31.6 & 28.7  \\
 & GZSL & 5.6 & 14.4 & 16.8   \\ 
\cmidrule(lr){2-5}
\multirow{2}{*}{\textbf{Our Approach}} & ZSL  & \textbf{26.3} & \textbf{33.1} & \textbf{30.7}     \\
 & GZSL & \textbf{9.3} &  \textbf{16.1}   & \textbf{19.0}     \\
\bottomrule[0.1em]
\end{tabular}
}
\vspace{-0.15cm}
\label{tab:sota_nuswide}
\end{table}

\noindent\textbf{Open Images:} Tab.~\ref{tab:sota_openimages} shows the state-of-the-art comparison for multi-label ZSL and GZSL tasks. The results are reported in terms of mAP and F1 score at \textit{top}-$K$ predictions ($K {\in} \{10,20\}$). We follow the same evaluation protocol as in the concurrent work of SDL~\cite{ben2021semantic}. Since Open Images has significantly larger number of labels, in comparison to NUS-WIDE, ranking them within an image is more challenging. This is reflected by the lower F1 scores in the table. Among existing methods, LESA obtains an mAP of $41.7\%$ for the ZSL task. In comparison, our approach outperforms LESA by achieving $73.6\%$ mAP with an absolute gain of $31.9\%$. Furthermore, our approach performs favorably against the best existing approach with F1 scores of $8.3$ and $5.5$ at $K {=} 10$ and $K {=} 20$.
It is worth noting that the ZSL task is challenging due to the high number of unseen labels ($400$). As in ZSL, our approach obtains a significant gain of $39.1\%$ mAP over the best published results for GZSL and also achieves favorable performance in F1. Additional details and results are presented in Appendix~\ref{sec:appn_quant_res}.

\begin{table}[t]
\centering
\caption{\textbf{State-of-the-art comparison for multi-label ZSL and GZSL tasks on Open Images}. Results are reported in terms of mAP and F1 score at $K{\in}\{10,20\}$. Our approach sets a new state of the art for both tasks, in terms of mAP and F1 score. Best results are in bold.}
\adjustbox{width=\linewidth}{
\begin{tabular}{ccccc} 
\toprule[0.15em]
\multirow{1}{*}{\textbf{ Method}} & \multirow{1}{*}{\textbf{Task }} & \multirow{1}{*}{\textbf{mAP }} & \multicolumn{1}{c}{\cellcolor[HTML]{EEEEEE}\textbf{F1 (K = 10) }} & \multicolumn{1}{c}{\cellcolor[HTML]{DAE8FC}\textbf{F1 (K = 20) }} \\
\toprule[0.15em]
\multirow{2}{*}{CONSE~\cite{norouzi2013zero}} & ZSL & 40.4 & 0.4 & 0.3   \\
 & GZSL & 43.5 &  2.6 & 2.4  \\ 
\cmidrule(lr){2-5}
\multirow{2}{*}{LabelEM~\cite{akata2015label}} & ZSL & 40.5 & 0.5  & 0.4   \\
 & GZSL & 45.2 &  5.2  & 5.1   \\ 
\cmidrule(lr){2-5}
\multirow{2}{*}{Fast0Tag~\cite{zhang2016fast}} & ZSL & 41.2 & 0.7 & 0.6  \\
 & GZSL & 45.2 & 16.0 & 12.9  \\ 
\cmidrule(lr){2-5}
\multirow{2}{*}{Attention per Cluster~\cite{huynh2020shared}} & ZSL & 40.7 & 1.2 &  0.9  \\
 & GZSL  & 44.9 & 16.9 & 13.5  \\ 
\cmidrule(lr){2-5}
\multirow{2}{*}{LESA~\cite{huynh2020shared}} & ZSL & 41.7 & 1.4 & 1.0  \\
 & GZSL & 45.4 & 17.4 & 14.3   \\ 
\cmidrule(lr){2-5}
\multirow{2}{*}{\textbf{Our Approach}} & ZSL  & \textbf{73.6} & \textbf{8.3} & \textbf{5.5}     \\
 & GZSL & \textbf{84.5} &  \textbf{19.1}   & \textbf{15.9}     \\
\bottomrule[0.1em]
\end{tabular}
}
\vspace{-0.2cm}
\label{tab:sota_openimages}
\end{table}
\subsection{Ablation Study}

\begin{figure}[t]
\centering
\includegraphics[clip=true, trim=2em 0em 0em 2em, width=0.9\columnwidth, keepaspectratio]{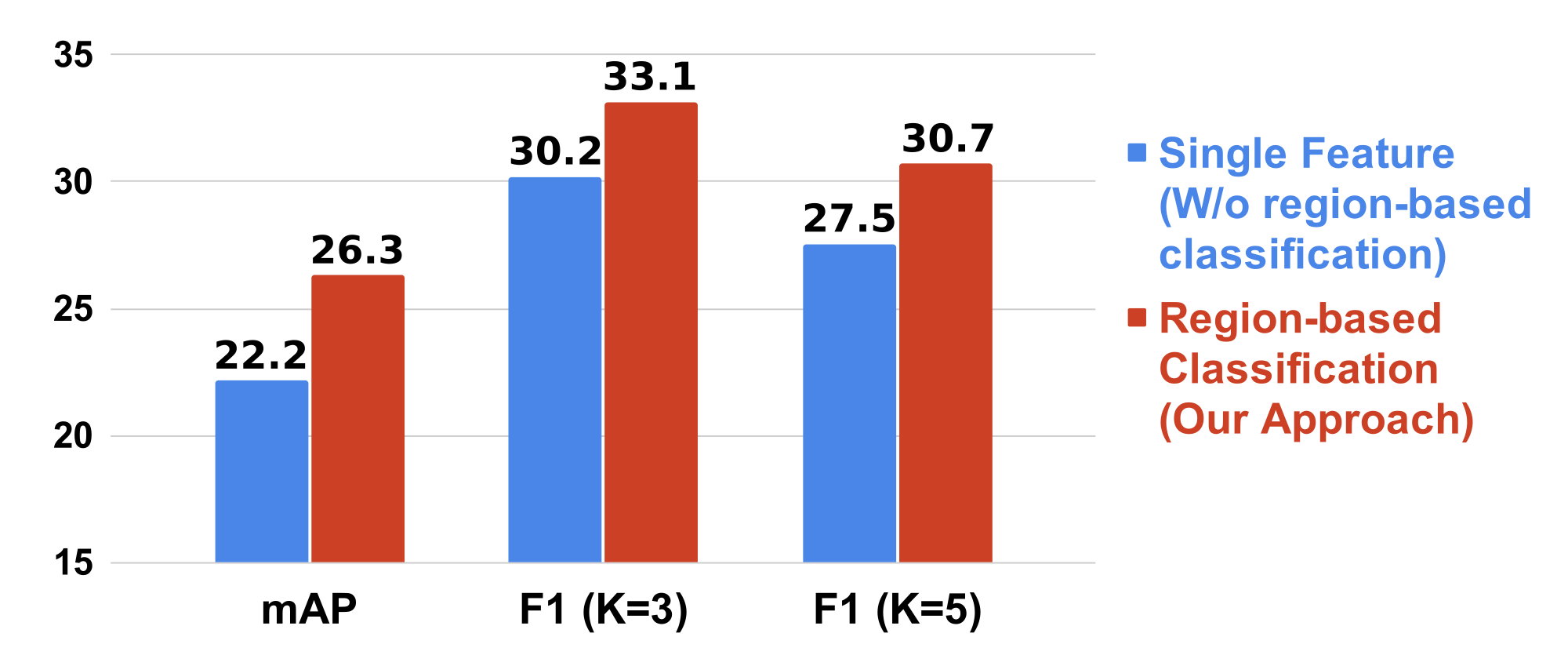}
\vspace{-0.2cm}
\caption{\textbf{Impact of region-based classification for the ZSL task on NUS-WIDE}, in terms of mAP and F1 at $K{\in}\{3,5\}$. Classifying spatially pooled features (blue bars) entangles the features of the different classes resulting in sub-optimal performance. In contrast, our proposed approach, which classifies each region individually and then spatially pools the \textit{per region} class scores (red bars), minimizes the inter-class feature entanglement and achieves superior classification performance.\vspace{-0.3cm}}
\label{fig:nuswide_regionwise}
\end{figure}
\begin{figure}[t]
    \centering
    \includegraphics[width=0.45\columnwidth]{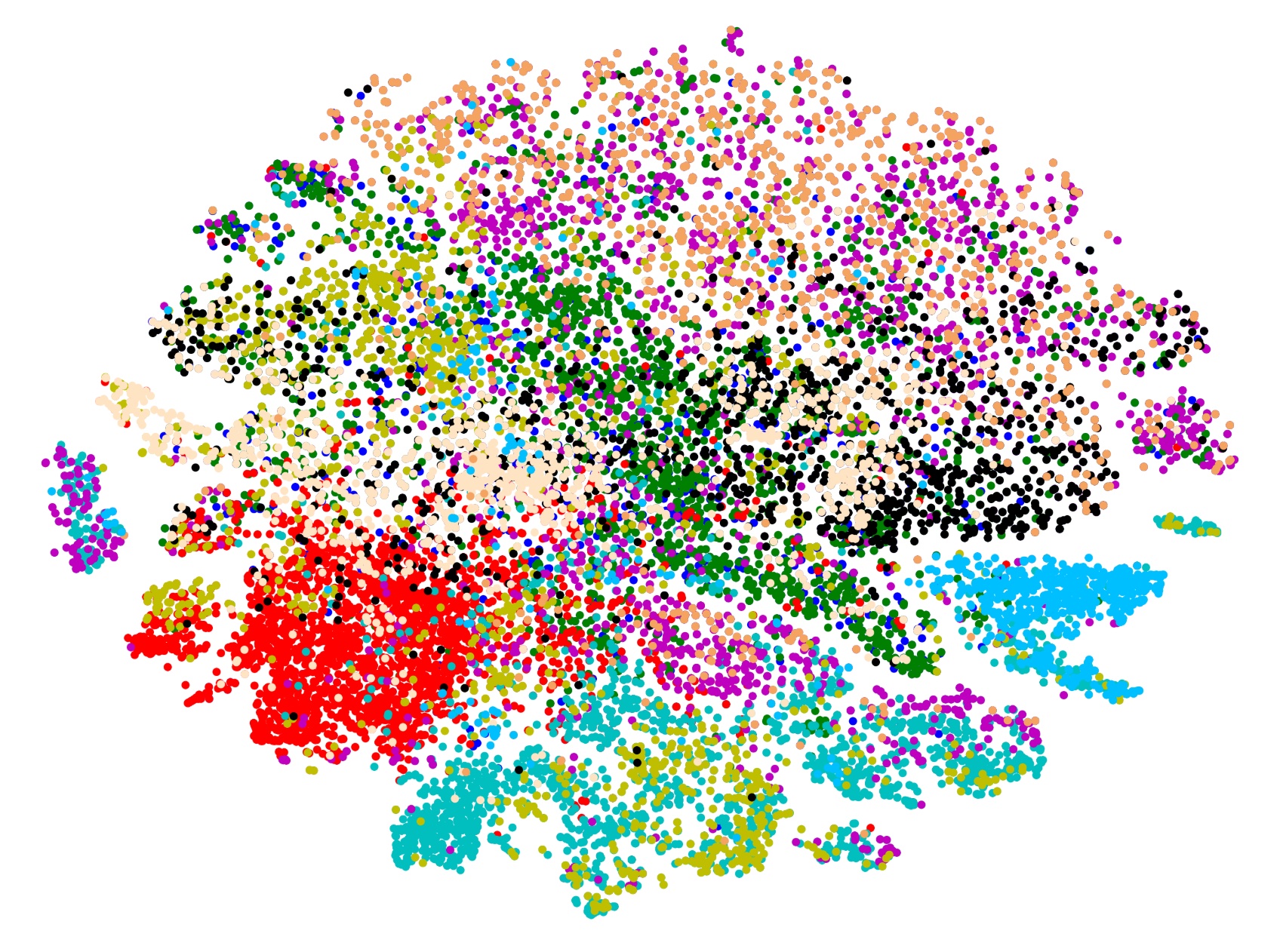}
    \includegraphics[width=0.45\columnwidth]{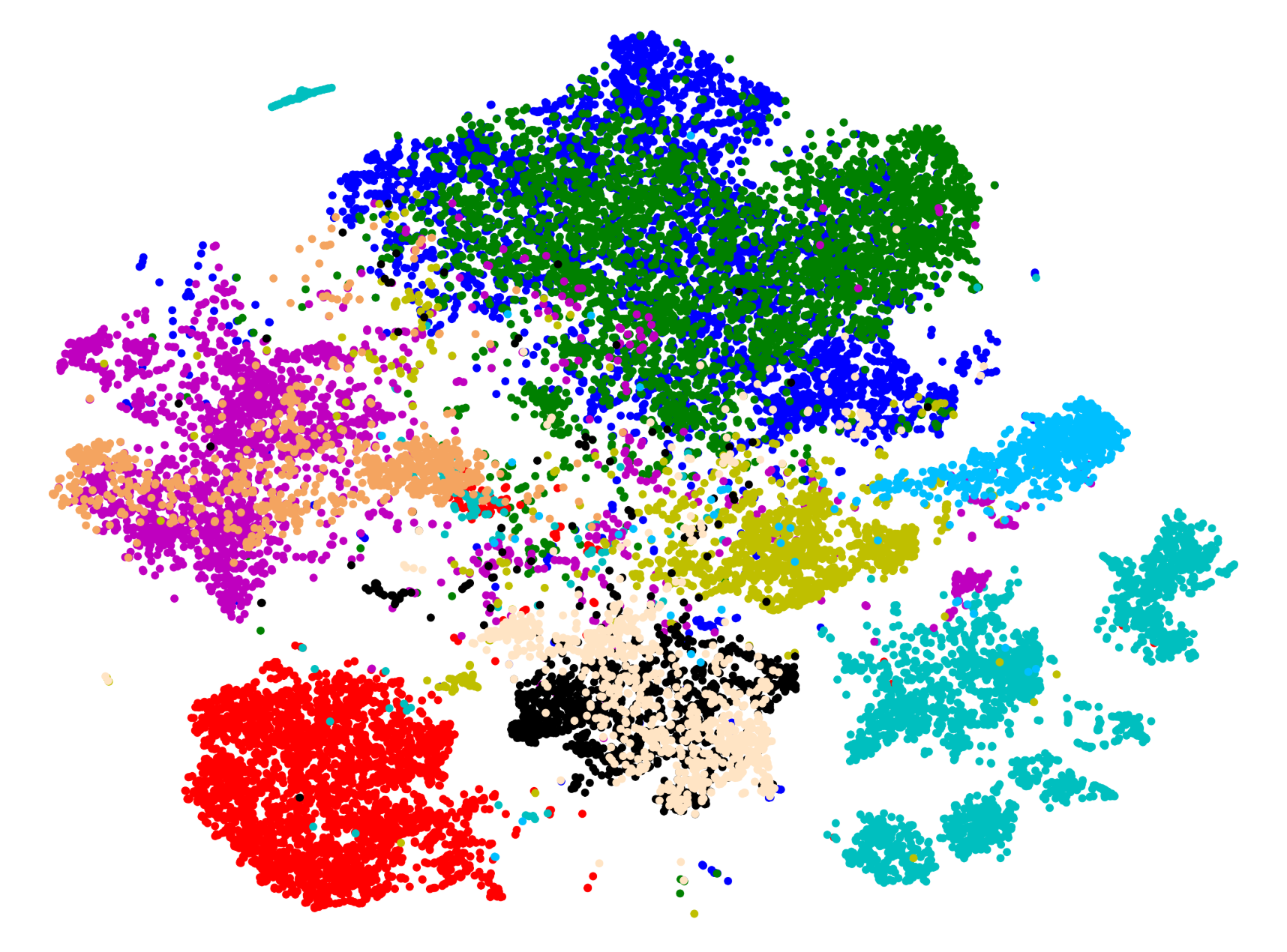}
    \caption{\textbf{t-SNE visualization showing the impact of the proposed region-level classification framework on the inter-class feature entanglement.} We present the comparison on $10$ unseen classes of NUS-WIDE. On left: the single feature representation-based classification pipeline, where the enriched features are spatially aggregated to obtain a feature vector (of length $d_r$) and then classified. On right: the proposed region-level classification framework, which classifies the region-level features first and then spatially pools the class scores to obtain image-level predictions. Our classification framework maintains the spatial resolution to preserve the region-level characteristics, thereby effectively minimizing the inter-class feature entanglement.\vspace{-0.2cm}
    }
    \label{fig:qual_tsne}
\end{figure}
\begin{figure}[t]
    \centering
    \includegraphics[width=\columnwidth]{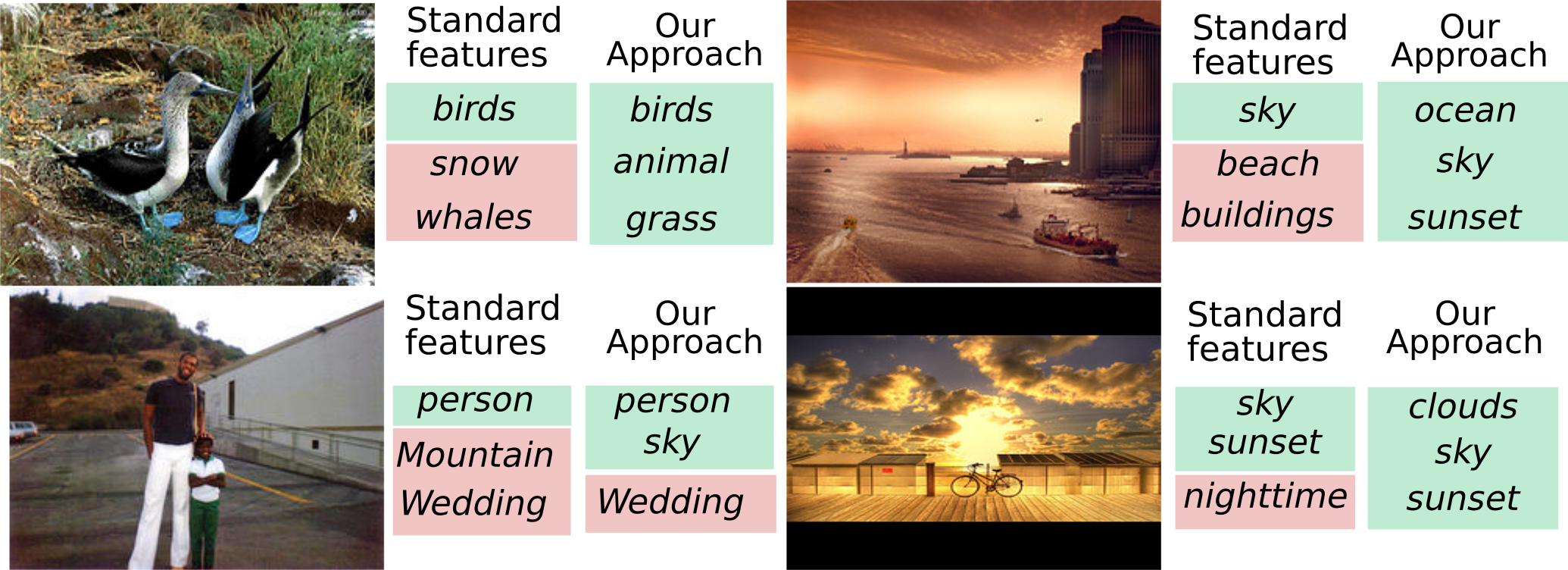}
    \caption{\textbf{Qualitative comparison on four test examples from NUS-WIDE, between the standard region features and our discriminative features}. \textit{Top}-$3$ predictions per image for both approaches are shown with \textcolor{mygreen}{true positives} and \textcolor{red}{false positives}. Compared to the standard region-based features, our approach learns discriminative region-based features and performs favorably.\vspace{-0.25cm}}
    \label{fig:nus_top3qual}
\end{figure}

\noindent\textbf{Impact of region-based classification:} 
To analyse this impact, we train our proposed framework without region-based classification, where the enriched features $\mathbf{e}_f$ are spatially average-pooled to a single feature representation (of size $d_r$) per image and then classified. Fig.~\ref{fig:nuswide_regionwise} shows the performance comparison between our frameworks trained with and without region-based classification in terms of mAP and F1. 
Since images have large and diverse set of positive labels, spatially aggregating features without the region-based classification (blue bars), leads to inter-class feature entanglement, as discussed in Sec.~\ref{sec:regionwise}. Instead, preserving the spatial dimension by classifying the region-based features, as in the proposed framework (red bars), mitigates the inter-class feature entanglement to a large extent. This leads to a superior performance for the region-based classification on both multi-label ZSL and GZSL tasks. These results suggest the importance of region-based classification for learning discriminative features in large-scale multi-label (G)ZSL tasks. 
Furthermore, Fig.~\ref{fig:qual_tsne} presents a t-SNE visualization showing the impact of our region-level classification framework on $10$ unseen classes from NUS-WIDE.
\begin{table}[t]
\centering
\caption{\textbf{Impact of the proposed BiAM comprising RCB and SCB blocks}. Note that all results here are reported with the same region-level classification framework and only the features utilized within the classification framework differs. Both RCB alone and SCB alone achieve consistently improved  performance over standard region features. For both ZSL and GZSL tasks, the best performance is obtained when utilizing the discriminative features obtained from the proposed BiAM. Best results are in bold.}
\setlength{\tabcolsep}{6pt}
\adjustbox{width=\linewidth}{
\begin{tabular}{ccccc} 
\toprule[0.15em]
\multirow{1}{*}{\textbf{ Method}} & \multirow{1}{*}{\textbf{Task }} & \multirow{1}{*}{\textbf{mAP }} & \multicolumn{1}{c}{\cellcolor[HTML]{EEEEEE}\textbf{F1 (K = 3) }} & \multicolumn{1}{c}{\cellcolor[HTML]{DAE8FC}\textbf{F1 (K = 5) }}   \\
\toprule[0.15em]
\multirow{1}{*}{Standard} & ZSL & 21.1 & 28.0  & 26.9   \\
\multirow{1}{*}{region features} & GZSL & 6.8 &  12.0  & 14.5   \\ 
\cmidrule(lr){2-5}
\multirow{2}{*}{RCB alone} & ZSL & 23.7 & 31.9 & 29.0  \\
 & GZSL & 7.6 & 14.7 & 17.6  \\ 
\cmidrule(lr){2-5}
\multirow{2}{*}{SCB alone} & ZSL  & 23.2 & 29.4 & 27.8 \\
 & GZSL & 8.6 & 14.0 & 16.7   \\ 
\cmidrule(lr){2-5}
\multirow{2}{*}{\textbf{BiAM (RCB + SCB)}} & ZSL  & \textbf{26.3} & \textbf{33.1} & \textbf{30.7}     \\
 & GZSL & \textbf{9.3} &  \textbf{16.1}   & \textbf{19.0}     \\
\bottomrule[0.1em]
\end{tabular}
}
\vspace{-0.2cm}
\label{tab:baseline_nuswide}
\end{table}
\begin{table}[t]
\caption{\label{tab:attn_variants}\textbf{ZSL comparison on NUS-WIDE with attention variants:} our attention (left) and other attentions~\cite{wang2018non,huang2019ccnet} (right).}
\begin{minipage}{0.5\linewidth}
        \adjustbox{width=\linewidth}{
            \begin{tabular}{cc} 
            \toprule[0.15em]
            \multirow{1}{*}{\textbf{Method}} & \multirow{1}{*}{\textbf{mAP}}\\
            \toprule[0.15em]
            BiAM: RCB w/ LayerNorm & 25.0 \\
            BiAM: RCB w/ \textit{sigmoid}  & 24.6 \\
            BiAM: SCB w/ \textit{softmax} & 24.3 \\
            \textbf{BiAM: Final} & \textbf{26.3}\\
            \bottomrule[0.1em]
            \end{tabular}%
            }
            \vspace{-0.25cm}
            \label{tab:biam_attn_ablations}
    \end{minipage}%
    \hspace{0.5cm}
        \begin{minipage}{0.4\linewidth}
        \centering
        \adjustbox{width=0.95\linewidth}{
        \begin{tabular}{cc} 
        \toprule[0.15em]
        \multirow{1}{*}{\textbf{ Method}} & \multirow{1}{*}{\textbf{mAP}}\\
        \toprule[0.15em]
        Non-Local~\cite{wang2018non} & 23.1 \\
        Criss-Cross Atn~\cite{huang2019ccnet} & 23.9  \\
        \textbf{BiAM (Ours)} & \textbf{26.3} \\
        \bottomrule[0.1em]
        \end{tabular}%
        }
        \vspace{-0.25cm}
        \label{tab:prev_attn_works}
\end{minipage}%
\end{table}%
\noindent\textbf{Impact of the proposed BiAM:}
Here, we analyse the impact of our feature enrichment mechanism (BiAM) to obtain discriminative feature representations. 
Tab.~\ref{tab:baseline_nuswide} presents the comparison between region-based classification pipelines based on standard features $\mathbf{h}_r$ and discriminative features $\mathbf{e}_f$ obtained from our BiAM on NUS-WIDE. We also present results of our RCB and SCB blocks alone. Both RCB alone and SCB alone consistently improve the (G)ZSL performance over the standard region-based features. This shows that our region-based classification pipeline benefits from the discriminative features obtained through the two complementary attention blocks. Furthermore, best results are obtained with our BiAM that comprises both RCB and SCB blocks, demonstrating the importance of encoding both region \textit{and} scene context information. 
Fig.~\ref{fig:nus_top3qual} shows a comparison between the standard features-based classification and the proposed classification framework utilizing BiAM on example unseen class images.

\begin{figure}[t]
    \centering
    \includegraphics[clip=true, trim=0em 0em 0em 1em,width=\columnwidth]{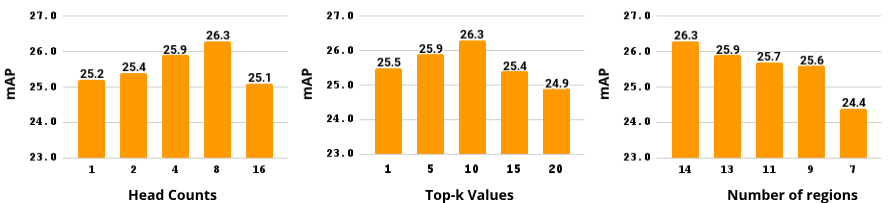}
    \caption{\textbf{ZSL comparison on NUS-WIDE when varying $H$, \textit{top}-$k$ and $h{\cdot}w$ regions}. Results improve slightly as heads $H$ increases till $8$ and drops beyond $8$, likely due to overfitting to seen classes. A similar trend is observed when \textit{top}-$k$ increases. Decreasing $h{\cdot}w$ regions from 14x14 to 9x9 does not affect much.\vspace{-0.5cm}}
    \label{fig:hyperparams}
\end{figure}

\noindent\textbf{Varying the attention modules:} 
Tab.~\ref{tab:attn_variants} (left) shows the comparison on NUS-WIDE when ablating RCB and SCB modules in our BiAM. Including LayerNorm in RCB or replacing its \textit{softmax} with \textit{sigmoid} or replacing \textit{sigmoid} with \textit{softmax} in SCB result in sub-optimal performance compared to our final BiAM. Similarly, replacing our BiAM with existing Non-Local~\cite{wang2018non} and Criss-cross~\cite{huang2019ccnet} attention blocks also results in reduced performance (see Tab.~\ref{tab:attn_variants} (right)). This shows the efficacy of BiAM, which integrates both region and holistic scene context.\\
\noindent\textbf{Varying the hyperparameters:}
Fig.~\ref{fig:hyperparams} shows the ZSL performance of our framework when varying heads $H$, $k$ in \textit{top}-$k$ and number of regions ($h{\cdot} w$). Performance improves as $H$ is increased till $8$ and drops beyond $8$, likely due to overfitting to seen classes. Similarly, as \textit{top}-$k$ increases beyond $10$, features of spatially-small classes entangle and reduce the discriminability. Furthermore, decreasing the regions leads to multiple classes overlapping in the same regions causing feature entanglement and performance drop.\\
\noindent\textbf{Compute and run-time complexity:} Tab.~\ref{tab:runtime} shows that our approach achieves significant performance gains of $6.7$\% and $31.3$\% over LESA with \textit{comparable} FLOPs, memory cost, training and inference run-times, on NUS-WIDE and Open Images, respectively. For a fair comparison, both methods are run on the same Tesla V100.\\
\indent Additional examples \wrt failure cases of our model such as confusing abstract classes (\eg, \textit{sunset} \vs \textit{sunrise}) and fine-grained classes are provided in Appendix~\ref{sec:appn_qual_res}.

\begin{table}[t]
\centering
\caption{\textbf{Comparison of our BiAM with LESA in terms of ZSL performance (mAP), train and inference time, FLOPs and memory cost on NUS-WIDE (NUS) and Open Images (OI).} Our BiAM achieves significant gain in performance with comparable compute and run-time complexity, over LESA.}
\adjustbox{width=\linewidth}{
\begin{tabular}{cccccc} 
\toprule[0.15em]
\multirow{1}{*}{\textbf{ Method}} & \multirow{1}{*}{\textbf{mAP (NUS / OI)}}  & \multirow{1}{*}{\textbf{Train (NUS / OI)}} &
\multirow{1}{*}{\textbf{Inference}} & \multirow{1}{*}{\textbf{FLOPs}}  & \multirow{1}{*}{\textbf{Memory}} \\
\toprule[0.15em]
LESA [10] & 19.4 / 41.7 &  9.1 hrs / 35 hrs & 1.4 ms & 0.46 G & 2.6 GB \\
\textbf{BiAM (Ours)} & 26.1 / 73.0 & 7.5 hrs / 26 hrs  & 2.3 ms & 0.59 G & 2.8 GB  \\
\bottomrule[0.1em]
\end{tabular}%
}
\vspace{-0.2cm}
\label{tab:runtime}
\end{table}

\subsection{Standard Multi-label Classification\label{sec:std_mll}}
In addition to multi-label (generalized) zero-shot classification, we evaluate our proposed region-based classification framework on the standard multi-label classification task. Here, image instances for all the labels are present in training. The state-of-the-art comparison for the standard multi-label classification on NUS-WIDE with $81$ human annotated labels is shown in Tab.~\ref{tab:mll_nuswide}. Among existing methods, the work of~\cite{kim2018bilinear} and LESA~\cite{huynh2020shared} achieve mAP scores of $32.6$ and $31.5$, respectively. Our approach outperforms all published methods and achieves a significant gain of $15.2\%$ mAP over the state of the art. Furthermore, our approach performs favorably against existing methods in terms of F1.

\section{Related Work}

Several works \cite{xian2018feature,shen2020invertible,liu2021near,Mandal19cvpr,zhang2021plug,narayan2020latent,zhang2020deep} have researched the conventional single-label ZSL problem.
In contrast, a few works~\cite{mensink2014costa,zhang2016fast,lee2018multi,huynh2020shared,gupta2021generative} have investigated the more challenging problem of multi-label ZSL. Mensink \etal \cite{mensink2014costa} propose an approach based on using co-occurrence statistics for multi-label ZSL. Zhang \etal \cite{zhang2016fast} introduce a method that utilizes linear mappings and non-linear deep networks to approximate principal direction from an input image. The work of~\cite{lee2018multi} investigates incorporating knowledge graphs to reason about relationships between multiple labels. 
Recently, Huynh and Elhamifar~\cite{huynh2020shared} introduce a shared attention-based multi-label ZSL approach, where the shared attentions are label-agnostic and are trained to focus on relevant foreground
regions by utilizing a formulation based on multiple loss terms.

Context is known to play a crucial role in several vision problems, such as object recognition~\cite{Olivacontext07,GalleguillosCVIU,Sanghyun2018context, Zhang2020context}. 
Studies~\cite{Geirhos2019context,Brendel2019context} have shown that deep convolutional networks-based visual recognition models implicitly rely on contextual information. Recently, self-attention models have achieved promising performance for
machine translation and natural language processing ~\cite{vaswani2017attention,Felix2019context,Zihang2019context,Jacob2019context}. This has inspired studies to investigate self-attention and related ideas for vision tasks, such as object recognition~\cite{Ramachandran2019context}, image synthesis~\cite{Zhang2019context} and video prediction~\cite{Xiaolong2018context}. Self-attention strives to learn the relationships between elements
of a sequence by estimating the relevance of one item to other items. Motivated by its success in several vision tasks, we introduce a multi-label zero-shot region-based classification approach that utilizes self-attention in the proposed bi-level attention module to reason about all regions together using pair-wise relations between these regions. To complement the self-attentive region features with the holistic scene context information, we integrate a global scene prior which enables us to enrich the region-level features with both region and scene context information.

\begin{table}[t]
\centering
\caption{\textbf{State-of-the-art performance comparison for the standard multi-label classification on NUS-WIDE}. The results are reported in terms of mAP and F1 score at $K{\in}\{3,5\}$. Our proposed approach achieves superior performance compared to existing methods, with gains as high as $15.2\%$ in terms of mAP. Best results are in bold.}
\adjustbox{width=\linewidth}{
\begin{tabular}{cccc} 
\toprule[0.15em]
\multirow{1}{*}{\textbf{ Method}} & \multirow{1}{*}{\textbf{mAP }} & \multicolumn{1}{c}{\cellcolor[HTML]{EEEEEE}\textbf{F1 (K = 3) }} & \multicolumn{1}{c}{\cellcolor[HTML]{DAE8FC}\textbf{F1 (K = 5) }}   \\
\toprule[0.15em]
WARP~\cite{gong2013deep} & 3.1 & 54.4 & 49.4  \\ 
WSABIE~\cite{weston2011wsabie} & 3.1 & 53.8 &  49.2 \\ 
Logistic~\cite{tsoumakas2007multi} & 21.6 & 51.1  & 46.1 \\ 
Fast0Tag~\cite{zhang2016fast} & 22.4 & 53.8 & 48.6 \\ 
CNN-RNN~\cite{wang2016cnn} & 28.3 & 55.2 & 50.8 \\ 
LESA~\cite{huynh2020shared} & 31.5 & 58.0 & 52.0 \\ 
Attention per Cluster~\cite{huynh2020shared} & 31.7 & 56.6 & 50.7 \\ 
Attention per Label~\cite{kim2018bilinear} & 32.6 & 56.8 & 51.3  \\ 
\textbf{Our Approach} & \textbf{47.8} & \textbf{59.6} & \textbf{53.4}  \\
\bottomrule[0.1em]
\end{tabular}%
}
\vspace{-0.2cm}
\label{tab:mll_nuswide}
\end{table}

\section{Conclusion} We proposed a region-based classification framework comprising a bi-level attention module for large-scale multi-label zero-shot learning. The proposed classification framework design preserves the spatial resolution of features to retain the multi-class information disentangled. This enables to effectively deal with large number of co-existing categories in an image. To contextualize and enrich the region features in our classification framework, we introduced a bi-level attention module that incorporates both region and scene context information, generating discriminative feature representations. Our simple but effective approach sets a new state of the art on two large-scale benchmarks and obtains absolute gains as high as $31.9\%$ ZSL mAP, compared to the best published results.

{\small
\bibliographystyle{ieee_fullname}
\bibliography{egbib}
}

\clearpage

\appendix

\section{Additional Quantitative Results\label{sec:appn_quant_res}}

\subsection{Standard Multi-Label Learning}
Similar to Sec.~\ref{sec:std_mll}, where we evaluate our approach for the standard multi-label classification on the NUS-WIDE dataset~\cite{nuswide}, here, we also evaluate on the large-scale Open Images dataset~\cite{openimages}. Tab.~\ref{tab:mll_openimages} shows the state-of-the-art comparison for the standard multi-label classification on Open Images. Here, $7,186$ classes are used for both training and evaluation. Test samples with missing labels for these $7,186$ classes are removed during evaluation, as in~\cite{huynh2020shared}. Due to significantly larger number of labels in Open Images, ranking the labels within an image is more challenging. This is reflected by the lower F1 scores in the table.
Among existing methods, Fast0Tag~\cite{zhang2016fast} and LESA~\cite{huynh2020shared} achieve an F1 score of $13.1$ and $14.5$ at $K{=}20$. Our approach achieves favorable performance against the existing approaches, achieving an F1 score of $17.3$ at $K{=}20$. The proposed approach also achieves superior performance in terms of mAP score, compared to existing methods and obtains an absolute gain of $35.6\%$ mAP over the best existing method.

\subsection{Robustness to Backbone Variation}
In Sec.~\ref{sec:experiments}, for a fair comparison with existing works such as Fast0Tag~\cite{zhang2016fast} and LESA~\cite{huynh2020shared}, we employed a pretrained VGG-19~\cite{simonyan2014very} as the backbone for extracting region-level and global-level features of images. However, such supervisedly pretrained backbone will not strictly conform with the zero-shot paradigm if there is any overlap between the unseen classes and the classes used for pretraining. To avoid using a supervisedly pre-trained network, we conduct an experiment by using the recent self-supervised DINO~\cite{caron2021emerging} ResNet-50 backbone trained on ImageNet \textit{without any labels}. Tab.~\ref{tab:dino_backbone} shows that our approach (BiAM) significantly outperforms LESA~\cite{huynh2020shared} even with a self-supervised pretrained backbone on both benchmarks: NUS-WIDE~\cite{nuswide} and Open Images~\cite{openimages}. Absolute gains as high as $6.9\%$ mAP are obtained for NUS-WIDE on the ZSL task. Similar favorable gains are also obtained for the GZSL task on both datasets. These results show that irrespective of the backbone used for extracting the image features, our BiAM approach performs favorably against existing methods, achieving significant gains across different datasets on both ZSL and GZSL tasks.

\begin{table}[t]
\centering
\caption{\textbf{State-of-the-art performance comparison for the standard multi-label classification on Open Images}. The results are reported in terms of mAP and F1 score at $K{\in}\{10,20\}$. In comparison to existing approaches, our approach achieves favorable performance in terms of both mAP and F1. Best results are in bold.}
\adjustbox{width=\linewidth}{
\begin{tabular}{cccc} 
\toprule[0.15em]
\multirow{1}{*}{\textbf{ Method}} & \multirow{1}{*}{\textbf{mAP }} & \multicolumn{1}{c}{\cellcolor[HTML]{EEEEEE}\textbf{F1 (K = 10) }} & \multicolumn{1}{c}{\cellcolor[HTML]{DAE8FC}\textbf{F1 (K = 20) }}   \\
\toprule[0.15em]
WARP~\cite{gong2013deep} & 46.0 & 7.7 & 7.4  \\ 
WSABIE~\cite{weston2011wsabie} & 47.2 & 2.2 & 2.2 \\ 
CNN-RNN~\cite{wang2016cnn} & 41.0 & 9.6 & 10.5 \\ 
Logistic~\cite{tsoumakas2007multi} & 49.4 & 13.3 & 11.8 \\   
Fast0Tag~\cite{zhang2016fast} & 45.4 & 16.2 & 13.1 \\ 
One Attention per Cluster~\cite{huynh2020shared} & 45.1 & 16.3 & 13.0 \\
LESA~\cite{huynh2020shared} & 45.6 & 17.8 & 14.5 \\ 
\textbf{Our Approach} & \textbf{85.0} & \textbf{20.4} & \textbf{17.3} \\
\bottomrule[0.1em]
\end{tabular}%
}
\vspace{-0.2cm}
\label{tab:mll_openimages}
\end{table}

\begin{table}[t]
\centering
\caption{\textbf{ZSL/GZSL performance comparison with LESA on NUS-WIDE and Open Images, when using the recent DINO ResNet-50 backbone} pretrained on ImageNet \textit{without any labels}. Our BiAM outperforms LESA~\cite{huynh2020shared} with a large margin on both datasets.}
\adjustbox{width=\linewidth}{
\begin{tabular}{cccc|cc} 
\toprule[0.15em]
\multirow{2}{*}{\textbf{Backbone}} & \multirow{2}{*}{\textbf{Task}} &
 \multicolumn{2}{c}{\textbf{NUS-WIDE (mAP)}} &  \multicolumn{2}{c}{\textbf{Open Images (mAP)}} \\
& & LESA & \textbf{BiAM (Ours)} & LESA & \textbf{BiAM (Ours)} \\
\toprule[0.15em]
 \multirow{2}{*}{\textbf{DINO ResNet-50}~\cite{caron2021emerging}} & ZSL & 20.5 & \textbf{27.4} & 41.9 & \textbf{74.0}\\
 & GZSL & 6.4 & \textbf{10.2} & 45.5 & \textbf{84.8}\\
\bottomrule[0.1em]
\end{tabular}%
}
\vspace{-0.3cm}
\label{tab:dino_backbone}
\end{table}

\section{Additional Qualitative Results\label{sec:appn_qual_res}}
\begin{figure*}[t]
    \centering
    \includegraphics[width=\textwidth]{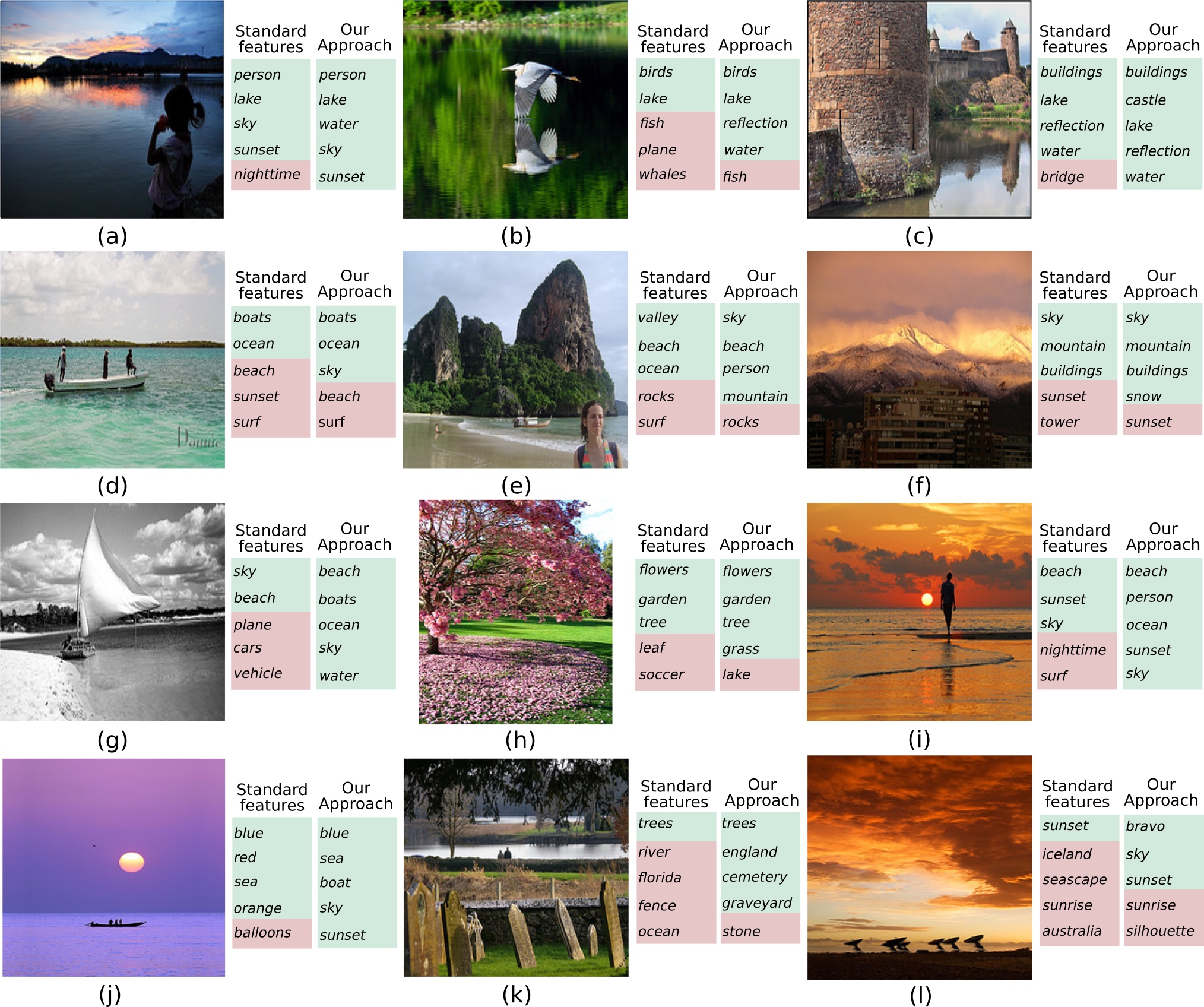}
    \caption{\textbf{Qualitative comparison for multi-label zero-shot classification on nine example images from the NUS-WIDE test set, between the standard region-based features and our discriminative features}. \textit{Top}-$5$ predictions per image for both approaches are shown with \textcolor{mygreen}{true positives} and \textcolor{red}{false positives}. 
    Generally, in comparison to the standard region-based features, our approach learns discriminative region-based features and results in increased true positive predictions along with reduced false positives. \Eg, \textit{reflection} and \textit{water} in (b), \textit{ocean} and \textit{sky} in (g), \textit{boat} and \textit{sky} in (j) along with \textit{graveyard} and \textit{england} in (k) are correctly predicted. Though a few confusing classes are predicted (\eg, \textit{beach} and \textit{surf} in (d)), the obvious false positives such as \textit{vehicle} in (g), \textit{soccer} in (h), \textit{balloons} in (j) and \textit{ocean} in (k) which are predicted by the standard region-based features, are reduced by our discriminative region-based features. These qualitative results suggest that our approach based on discriminative region features achieves promising performance in comparison to the standard features, for the task of multi-label (generalized) zero-shot classification.}
    \label{fig:zsl_classification}
\end{figure*}

\paragraph{Multi-label zero-shot classification:} Fig.~\ref{fig:zsl_classification} shows the qualitative results for multi-label (generalized) zero-shot learning. Nine example images from the test set of the NUS-WIDE dataset~\cite{nuswide} are presented in each figure. The comparison is shown between the standard region-based features and our discriminative region-based features. Alongside each image, \textit{top}-$5$ predictions for both approaches are shown with \textcolor{mygreen}{true positives} and \textcolor{red}{false positives}. In general, our approach learns discriminative region-based features and achieves increased true positive predictions along with reduced false positives, compared to the standard region-based features.
\Eg, categories such as \textit{reflection} and \textit{water} in Fig.~\ref{fig:zsl_classification}(b), \textit{ocean} and \textit{sky} in Fig.~\ref{fig:zsl_classification}(g), \textit{boat} and \textit{sky} in Fig.~\ref{fig:zsl_classification}(j) along with \textit{graveyard} and \textit{england} in Fig.~\ref{fig:zsl_classification}(k) are correctly predicted. Both approaches predict a few confusing classes such as \textit{beach} and \textit{surf} in Fig.~\ref{fig:zsl_classification}(d) in addition to \textit{sunrise} and \textit{sunset} that are hard to differentiate using visual cues alone in Fig.~\ref{fig:zsl_classification}(l). Moreover, false positives that are predicted by the standard region-based features, are reduced by our discriminative region-based features, \eg, \textit{vehicle} in Fig.~\ref{fig:zsl_classification}(g), \textit{soccer} in Fig.~\ref{fig:zsl_classification}(h), \textit{balloons} in Fig.~\ref{fig:zsl_classification}(j), and \textit{ocean} in Fig.~\ref{fig:zsl_classification}(k). These results suggest that our approach based on discriminative region features achieves promising performance against the standard features, for multi-label (generalized) zero-shot classification.

\begin{figure*}[t]
    \centering
    \includegraphics[width=\textwidth]{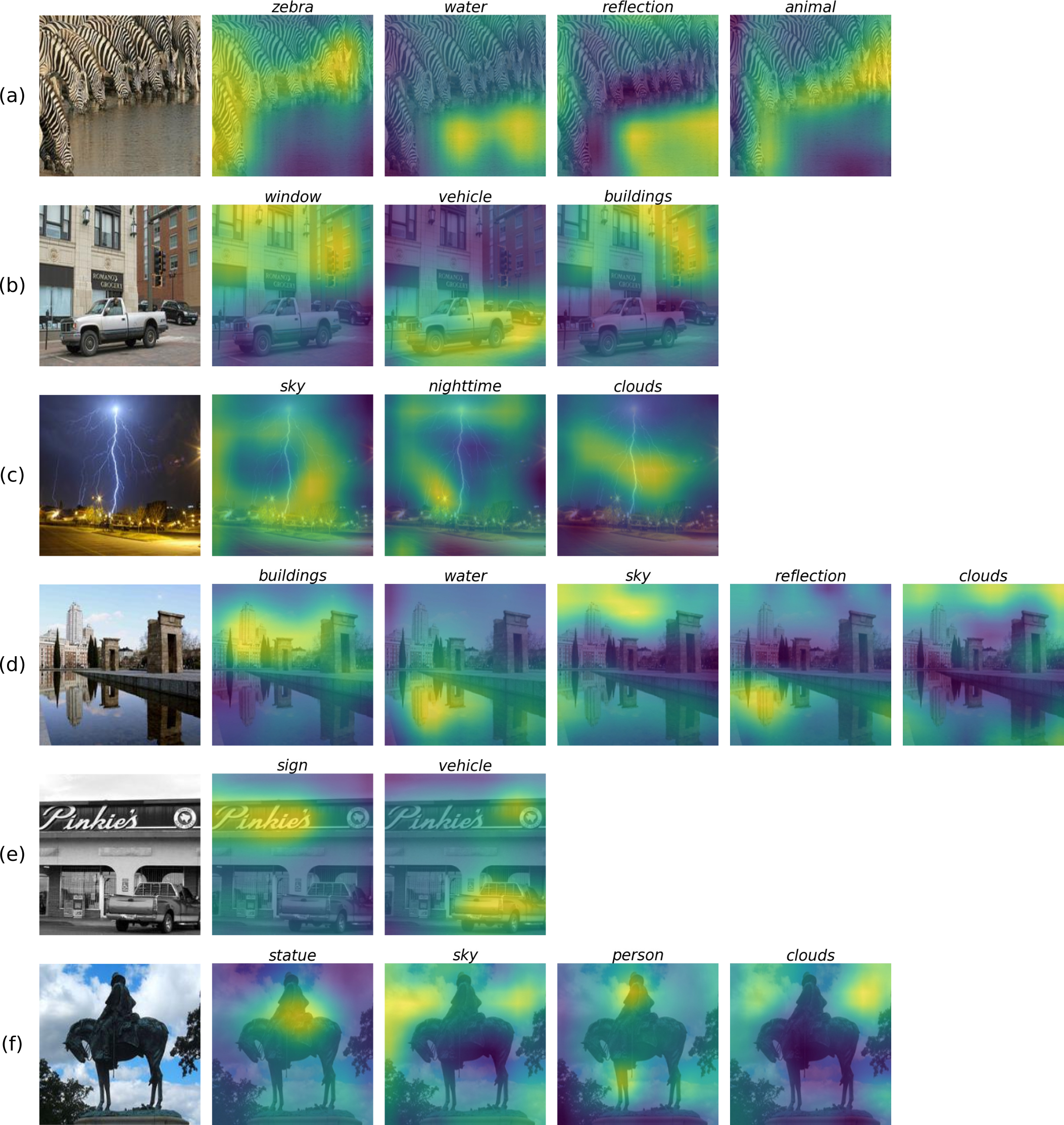}
    \caption{\textbf{Qualitative results with attention maps generated by our proposed approach, on example test images from the NUS-WIDE~\cite{nuswide} dataset}. For each image, class-specific maps for the ground truth unseen classes are shown with the corresponding labels on top. Generally, we observe that these maps focus reasonably well on the desired  classes. \Eg, promising attention/focus is observed on classes such as \textit{zebra} in (a), \textit{vehicle} in (b), \textit{buildings} in (d) and \textit{statue} in (f). Although we observe that the attention maps of visually similar classes such as \textit{sky} and \textit{clouds} overlap, as in (d), these abstract classes, including \textit{reflection} in (a), (d) and \textit{nighttime} in (c) are well captured. These qualitative results show that our proposed approach generates promising class-specific attention maps, leading to improved multi-label (generalized) zero-shot classification.}
    \label{fig:nuswide_qual}
\end{figure*}

\begin{figure*}[t]
    \centering
    \includegraphics[width=\textwidth]{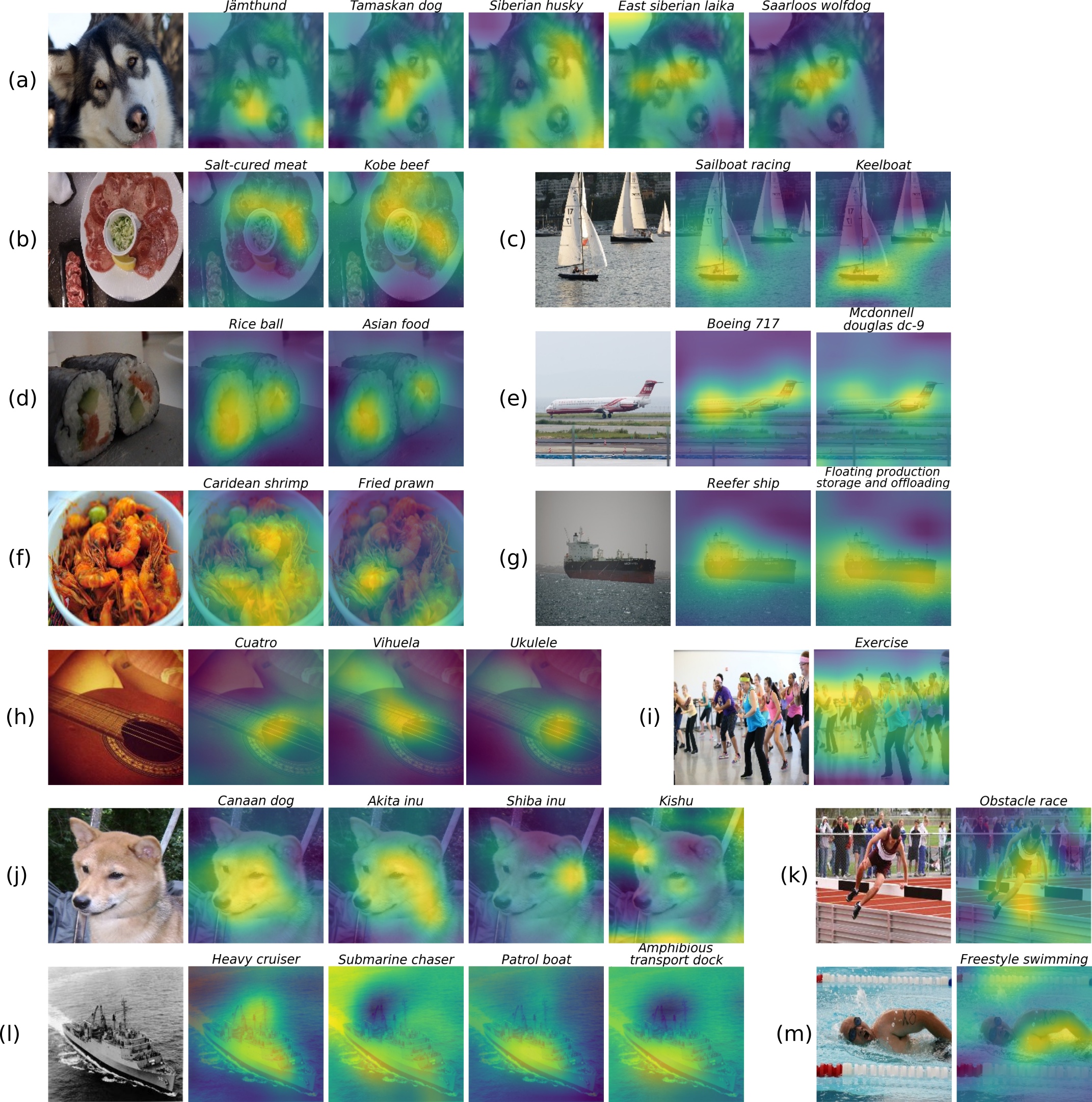}
    \caption{\textbf{Qualitative results with attention maps generated by our proposed approach, on example test images from the Open Images~\cite{openimages} dataset}. For each image, class-specific maps for the ground truth unseen classes are shown with the corresponding labels on top. Although there are overlapping attention regions for visually similar and fine-grained classes (\eg, \textit{Caridean shrimp} and \textit{Fried prawn} in (f), \textit{Canaan dog} and \textit{Akita inu} in (j)),  generally, these maps focus reasonably well on the desired classes. \Eg, promising class-specific attention is captured for \textit{Keelboat} in (c), \textit{Boeing 717} in (e) and \textit{Exercise} in (i). These qualitative results show that our proposed approach generates promising class-specific attention maps, resulting in improved multi-label (generalized) zero-shot classification.}
    \label{fig:openimages_qual}
\end{figure*}

\paragraph{Visualization of attention maps:} Fig.~\ref{fig:nuswide_qual} and~\ref{fig:openimages_qual} show the visualizations of attention maps for the ground truth classes in example test images from NUS-WIDE and Open Images, respectively. Alongside each example, class-specific maps for the unseen classes are shown with the corresponding labels on top. In general, we observe that these maps focus reasonably well on the desired classes. \Eg, promising class-specific attention is captured for \textit{zebra} in Fig.~\ref{fig:nuswide_qual}(a), \textit{vehicle} in Fig.~\ref{fig:nuswide_qual}(b), \textit{buildings} in Fig.~\ref{fig:nuswide_qual}(d), \textit{Keelboat} in Fig.~\ref{fig:openimages_qual}(c), \textit{Boeing 717} in Fig.~\ref{fig:openimages_qual}(e) and \textit{Exercise} in Fig.~\ref{fig:openimages_qual}(i). Although we observe that the attention maps of visually similar classes overlap for \textit{sky} and \textit{clouds} in Fig.~\ref{fig:nuswide_qual}(d), these abstract categories, including \textit{reflection} in Fig.~\ref{fig:nuswide_qual}(a) and \textit{nighttime} in Fig.~\ref{fig:nuswide_qual}(c) are well captured. These qualitative results show that our proposed approach (BiAM) generates promising class-specific attention maps, leading to improved multi-label (generalized) zero-shot classification.

\end{document}